\documentclass[conference]{IEEEtran}
\IEEEoverridecommandlockouts
\usepackage{cite}
\usepackage{url}
\usepackage[T1]{fontenc}
\usepackage{times}
\usepackage{amsmath,amssymb,amsfonts,amsthm}
\usepackage{algorithmic}
\usepackage{graphicx}
\usepackage{textcomp}
\usepackage{xcolor}
\def\BibTeX{{\rm B\kern-.05em{\sc i\kern-.025em b}\kern-.08em
    T\kern-.1667em\lower.7ex\hbox{E}\kern-.125emX}}

\usepackage{graphicx}
\usepackage{multirow}
\usepackage{subfigure}
\usepackage{paralist}
\usepackage{balance} 
\usepackage{microtype}
\usepackage{amsmath, amssymb}

\usepackage{color}

\usepackage{amsmath,amsfonts,bm}
\usepackage{mathtools}

\def\eqref#1{equation~\ref{#1}}

\def\1{\bm{1}}

\def\rvh{{\mathbf{h}}}

\def\rvo{{\mathbf{o}}}

\def\rvr{{\mathbf{r}}}

\def\rvv{{\mathbf{v}}}
\def\rvw{{\mathbf{w}}}
\def\rvx{{\mathbf{x}}}

\def\rmA{{\mathbf{A}}}
\def\rmB{{\mathbf{B}}}

\def\rmK{{\mathbf{K}}}

\def\rmO{{\mathbf{O}}}

\def\rmQ{{\mathbf{Q}}}

\def\rmV{{\mathbf{V}}}
\def\rmW{{\mathbf{W}}}

\DeclareMathAlphabet{\mathsfit}{\encodingdefault}{\sfdefault}{m}{sl}
\SetMathAlphabet{\mathsfit}{bold}{\encodingdefault}{\sfdefault}{bx}{n}

\def\gD{{\mathcal{D}}}

\def\gL{{\mathcal{L}}}

\def\gT{{\mathcal{T}}}
\def\gU{{\mathcal{U}}}
\def\gV{{\mathcal{V}}}

\def\gX{{\mathcal{X}}}

\def\sR{{\mathbb{R}}}

\newcommand{\softmax}{\mathrm{softmax}}

\newcommand{\Oplus}{\ensuremath{\vcenter{\hbox{\scalebox{1.5}{$\oplus$}}}}}
\newcommand{\Concat}{\Oplus}

\DeclareMathOperator*{\argmin}{arg\,min}

\newcommand{\defeq}{\vcentcolon=}

\newtheorem{remark}{Remark}

\usepackage{color}

\newcommand{\red}[1]{{\color{red}{#1}}}

\newtheorem{theorem}{Theorem}

\def \BASE {\mbox{BASE}}
\def \BERT {\mbox{BERT}}

\def \BERTBASE {\mbox{BERT}_{\mbox{\tiny BASE}}}
\def \BERTLARGE {\mbox{BERT}_{\mbox{\tiny LARGE}}}
\def \softmax {\mbox{softmax}}
\def \Att {\mbox{Att}}
\def \MH {\mbox{MH}}
\def \head {\mbox{head}}

\def \LN {{\rm LN}}
\def \FFN {{\rm FFN}}

\def \MSE {{\rm MSE}}

\begin{document}

\title{RefBERT: Compressing BERT by Referencing to Pre-computed Representations}

\author{\IEEEauthorblockN{Xinyi Wang$^1$\textsuperscript{\textsection},
Haiqin Yang$^2$\textsuperscript{\textsection}, Liang Zhao$^2$, Yang Mo$^2$, and
Jianping Shen$^2$}
\IEEEauthorblockA{$^1${University of California, Santa Barbara} USA \\
$^2$Ping An Life Insurance Company of China, Ltd., Shenzhen, China\\
Email: $^1$xinyi\_wang@ucsb.edu,
$^2$hqyang@ieee.org, $^2$\{moyang853, zhaoliang425,  shenjianping324\}@pingan.com.cn
}
}
\maketitle
\begingroup\renewcommand\thefootnote{\textsection}
\footnotetext{Authors contribute equally.  The work was partially done when Xinyi was an intern at Ping An Life.  Haiqin Yang is the corresponding author.}
\endgroup
\begin{abstract}
Recently developed large pre-trained language models, e.g., BERT, have achieved remarkable performance in many downstream natural language processing applications.  These pre-trained language models often contain hundreds of millions of parameters and suffer from high computation and latency in real-world applications.  It is desirable to reduce the computation overhead of the models for fast training and inference while keeping the model performance in downstream applications.  Several lines of work utilize knowledge distillation to compress the teacher model to a smaller student model.  However, they usually discard the teacher's knowledge when in inference.  Differently, in this paper, we propose RefBERT to leverage the knowledge learned from the teacher, i.e., facilitating the pre-computed BERT representation on the reference sample and compressing BERT into a smaller student model.  To guarantee our proposal, we provide theoretical justification on the loss function and the usage of reference samples.  Significantly, the theoretical result shows that including the pre-computed teacher's representations on the reference samples indeed increases the mutual information in learning the student model.  Finally, we conduct the empirical evaluation and show that our RefBERT can beat the vanilla TinyBERT over 8.1\% and achieves more than 94\% of the performance of $\BERTBASE$ on the GLUE benchmark.   Meanwhile, RefBERT is 7.4x smaller and 9.5x faster on inference than $\BERTBASE$. 
\end{abstract}
\begin{IEEEkeywords}
BERT, knowledge distillation, reference, mutual information
\end{IEEEkeywords}

\section{Introduction}

Recently, the NLP community has witnessed the effectiveness of utilizing large pre-trained language models, such as BERT~\cite{DBLP:conf/naacl/DevlinCLT19}, ELECTRA~\cite{DBLP:conf/iclr/ClarkLLM20}, XLNet~\cite{DBLP:conf/nips/YangDYCSL19}, and open-GPT~\cite{DBLP:journals/crossroads/CohenG20}.  These large language models have significantly improved the performance in many NLP downstream tasks, e.g., the GLUE benchmark~\cite{DBLP:conf/iclr/WangSMHLB19}.  However, the pre-trained language models usually contain hundreds of millions of parameters and suffer from high computation and latency in real-world applications~\cite{lei2017swim,conf/IJCNN/YangS21,DBLP:journals/corr/abs-2103-08886}.  Hence, in order to make the large pre-trained language models applicable for broader applications~\cite{gao2021advances,zhu2021retrieving,DBLP:journals/jbi/HuZYCZ17,DBLP:conf/naacl/JiaoYKL19}, it is necessary to reduce the computation overhead to accelerate the finetuning and inference of the large pre-trained language models.

In the literature, several effective techniques, such as quantization~\cite{DBLP:journals/corr/HanMD15,DBLP:conf/nips/HouZKGQL19} and knowledge distillation~\cite{DBLP:conf/emnlp/SunCGL19,DBLP:journals/corr/abs-1910-01108,DBLP:journals/corr/abs-1909-10351}, have been explored to tackle the computation overhead in large pre-trained language models.  In this paper, we focus on knowledge distillation~\cite{DBLP:journals/corr/HintonVD15} to transfer the knowledge embedded in a large teacher model to a small student model because knowledge distillation has been proved its effectiveness in several pieces of work, e.g., DistilBERT~\cite{DBLP:journals/corr/abs-1910-01108} and TinyBERT~\cite{DBLP:journals/corr/abs-1909-10351}.  However, these methods usually discard the teacher's representations after obtaining a learned student model, which may yield a sub-optimal performance. 

Different from existing work, we propose RefBERT to utilize the teacher's representations on reference samples to compress BERT into a small student model while maintaining the model performance.  Here, a reference sample of an input sample refers to the most similar (but not the same one) in the dataset evaluated by a certain similarity criterion, e.g., containing the most common keywords or following the most similar structure.  Our RefBERT delivers two modifications on the original Transformer layers: (1) {\bf the key and the value in the first Transformer layer:} the key in the multi-head attention networks of the first Transformer layer of the student model will facilitate both the student's embedding on the input sample and the teacher's embedding on the reference sample while the corresponding value in the multi-head attention networks will utilize both the student's embedding on the input sample and the teacher's last layer's representation on the reference sample.  By simply concatenating them, we can effectively absorb their information through the self-attention mechanism.  (2) {\bf Shifting the normalization attention score:} we subtract the attention score (normalized by the softmax function) a constant to amplify the effect of the normalization attention score.  By subtracting a constant in the score, we can place negative attention for non-informative tokens and discard the impact of unrelated parts in the next layer.  We then follow the setup of TinyBERT to learn the student's parameters by distilling the embedding-layer, hidden states, attention weights, and the prediction layer.  More importantly, we present theoretical analysis to justify the selection of the mean-square-error (MSE) loss function while revealing that by including any related reference sample during the compression procedure, our RefBERT indeed increases the information absorption. 

We highlight the contribution of our work as follows: 
\begin{compactitem}
\item We propose a novel knowledge distillation method, namely RefBERT, for Transformer-based architecture to transfer the linguistic knowledge encoded in the teacher BERT to a small student model through the reference mechanism.  Hence, the teacher's information will be facilitated by the student model during inference.
\item We modify the query, key, and value of the multi-head attention networks at the first layer of the student model to facilitate the teacher's embeddings and the representations in the last Transformer layer.  The attention score is subtracted by a constant to discard the effect of non-informative tokens.  More importantly, theoretical analysis has provided to justify the selection of the MSE loss and the including of reference samples in information absorption. 
\item We conduct experimental evaluations and show that our RefBERT can beat TinyBERT without data augmentation over 8.1\% and attains more than 94\% the performance of the teacher $\BERTBASE$ on the GLUE benchmark.  Meanwhile, RefBERT is 7.4x smaller and 9.5x faster on inference than $\BERTBASE$.  
\end{compactitem}


{
}













\section{Related Work}
We review two main streams of related work in the following:

{
{\bf Pre-training.}  In natural language processing, researchers have trained models on huge unlabeled text to learn precise word representations.  The models range from word embeddings~\cite{DBLP:conf/nips/MikolovSCCD13,DBLP:conf/emnlp/PenningtonSM14} to contextual word representations~\cite{DBLP:conf/naacl/PetersNIGCLZ18,liang2020pirhdy} and recently-developed powerful pre-trained language models, such as BERT~\cite{DBLP:conf/naacl/DevlinCLT19}, ELECTRA~\cite{DBLP:conf/iclr/ClarkLLM20}, XLNet~\cite{DBLP:conf/nips/YangDYCSL19}, and open-GPT~\cite{DBLP:journals/crossroads/CohenG20}.  After that, researchers usually apply the fine-tuning mechanism to update the large pre-trained representations for downstream  tasks with a small number of task-specific parameters~\cite{DBLP:conf/naacl/DevlinCLT19,DBLP:journals/corr/abs-1907-11692,DBLP:journals/corr/abs-1909-11942,lei2021have}.  However, the high demand for computational resources has hindered the applicability to a broader range, especially those resource-limited applications.

{\bf Distillation with unsupervised pre-training.} Early attempts try to leverage the unsupervised pre-training representations for behaviors distillation~\cite{DBLP:conf/naacl/PetersNIGCLZ18}, e.g., label distillation~\cite{DBLP:journals/corr/abs-1909-03508,DBLP:conf/emnlp/SunCGL19} and task-specific knowledge distillation~\cite{DBLP:journals/corr/abs-1903-12136}.  The distillation procedure is also extended from single-task learning to multi-task learning, i.e., distilling knowledge from multiple teacher models to a light-weight student model~\cite{DBLP:conf/acl/ClarkLKML19,DBLP:conf/wsdm/YangSGLJ20}.  Knowledge distillation~\cite{DBLP:journals/corr/HintonVD15} has been frequently applied to compress a larger teacher model to a compact student model~\cite{DBLP:journals/corr/abs-1908-08962}.  For example, DistilBERT~\cite{DBLP:journals/corr/abs-1910-01108} attempts to compress the intermediate layers of BERT into a smaller student model.  TinyBERT~\cite{DBLP:journals/corr/abs-1909-10351} further explores more distillation losses while facilitating data augmentation to attain good performance for downstream NLP tasks.  InfoBERT~\cite{DBLP:journals/corr/abs-2010-02329} tries to improve the model robustness based on  information-theoretic guarantee.  However, these methods discard the teacher's direct information after obtaining the student model.  This may lack sufficient information and yield sub-optimal performance during inference. 





}

\section{Notations and Problem Statement}
We first define some notations.  Bold capital letters, e.g., $\rmK$, indicate matrices.  Bold small letters, e.g., $\rvw$, indicate vectors or sequences. Letters in calligraphic or blackboard bold fonts, e.g., $\gX$, $\sR$, indicate a set, where $\sR^n$ denotes an $n$-dimensional real space.   With a little abuse of notations, we use lowercase letters to denote indices or the density functions (e.g., $p$, $q$), while uppercase letters for probability measure (e.g., $P$, $Q$), or random variables.  $\top$ denotes the transpose operator.  $\Concat$ denotes the concatenation of vectors.  $\MSE$ defines the mean squared error loss function.  

The task of knowledge distillation from a {\bf teacher} model is to build a compact {\bf students} model to fit the real-world constraints, e.g., memory and latency budget, in downstream tasks.  In this paper, we fix the model architecture to Transformer because it is a well-known architecture attaining superior performance in a wide range of NLP tasks~\cite{DBLP:conf/nips/VaswaniSPUJGKP17,DBLP:conf/naacl/DevlinCLT19}.   

To relieve the burden of the model expression, we define the vanilla Transformer layer for the encoder ($\gT$) as follows: 
\begin{align}\label{eq:Tencoder} 
     \rvh_i &=\gT(\rvh_{i-1})\defeq\left\{
    \begin{array}{@{}l@{}l@{}}
    \rmA &=\MH(\rvh_{i-1}, \rvh_{i-1}, \rvh_{i-1}), \\
    \rmB &= \LN(\rvh_{i-1}+\rmA),\\
    \rvh_i &= \LN(\FFN(\rmB)+\rmB),
    \end{array}
    \right.
\end{align}%
where the function $\FFN$ consists of two linear transformations with an activation, e.g., GeLU or ReLU~\cite{DBLP:conf/nips/VaswaniSPUJGKP17}.  The multi-head attention ($\MH$) is computed by 
\begin{align}\label{eq:MH}
&\MH(\rmQ, \rmK, \rmV) = \Oplus_{h=1}^H \head_h\rmW^O, \\\label{eq:head_i}
&\head_h = \Att(\rmQ \rmW^h_Q, \rmK \rmW^h_K, \rmV \rmW^h_V),~h=1, \ldots,  H,\\\label{eq:att_layer}
&\Att(\rmQ, \rmK, \rmV)=\softmax\left(Att(\rmQ, \rmK)\right)\rmV,\\\label{eq:att}
& Att(\rmQ, \rmK) = \frac{\rmQ\rmK^\top}{\sqrt{d_H}},
\end{align}
where $H$ is the number of attention heads, $d_H$ is the teacher' hidden size.   

Hence, for an input $\rvx$ with $|\rvx|$ sub-words, i.e., $\rvx=x^{(1)}x^{(2)}\ldots x^{(|\rvx|)}$, we have $\rvh_l \in \sR^{|\rvx|\times d_H}$ and is computed by 
\begin{align}\label{eq:h_l}
    \rvh_l = \left\{\begin{array}{@{}ll} 
   Emb(\rvx), & l=0  \\
 \gT(\rvh_{l-1}), & l=1, \ldots, L, 
 \end{array}\right.
\end{align}
where $Emb(\cdot)$ is the embedding of the teacher model, which is usually computed by the summation of token embedding, position embedding, and segment embedding if there is.  In the compression, we will utilize the teacher information, $Emb(\cdot)$ and $Hid(\cdot)=\rvh_L$ (the output in the last Transformer layer), to derive the corresponding student model. 

For fast experimentation, we use a single teacher, without making a statement about the best architectural choice, and adopt the pre-trained $\BERTBASE$~\cite{DBLP:conf/naacl/DevlinCLT19} as the teacher model.  Other models, e.g., $\BERTLARGE$, XLNet~\cite{DBLP:conf/nips/YangDYCSL19}, can be easily deployed and tested within our framework.  To lessen the variation of the model, we also apply the same number of self-attention heads to both the teacher and the student models.

We are given the following data: 
\begin{compactitem}
\item {\bf Unlabeled language model data} ($\gD_{LM}$): a collection of texts for representation learning.  These corpora are usually collected from Wiki or other open domains, which contain thousands of millions of words without strong domain information.   
\item {\bf Labeled data} ($\gD_{L}$): a set of $N$ training examples $\{(\rvx_i, y_i)\}_{i=1}^N$, where $\rvx_i\in\sR^d\subseteq\gX$ is an input in a $d$-dimensional space of $\gX$ and $y_i$ is the response.  For most NLP downstream tasks, labeled data requires a lot of experts' manual effort and are thus restricted in size.
\end{compactitem}
In our work, we apply $\gD_{LM}$ to train the student model from the teacher model as stated in the next section. 

\section{Methodology}

In this section, we present our proposed RefBERT and provide the theoretical justification to support our proposal.

\begin{figure*}[t]
    \centering
    \includegraphics[width=0.8\textwidth]{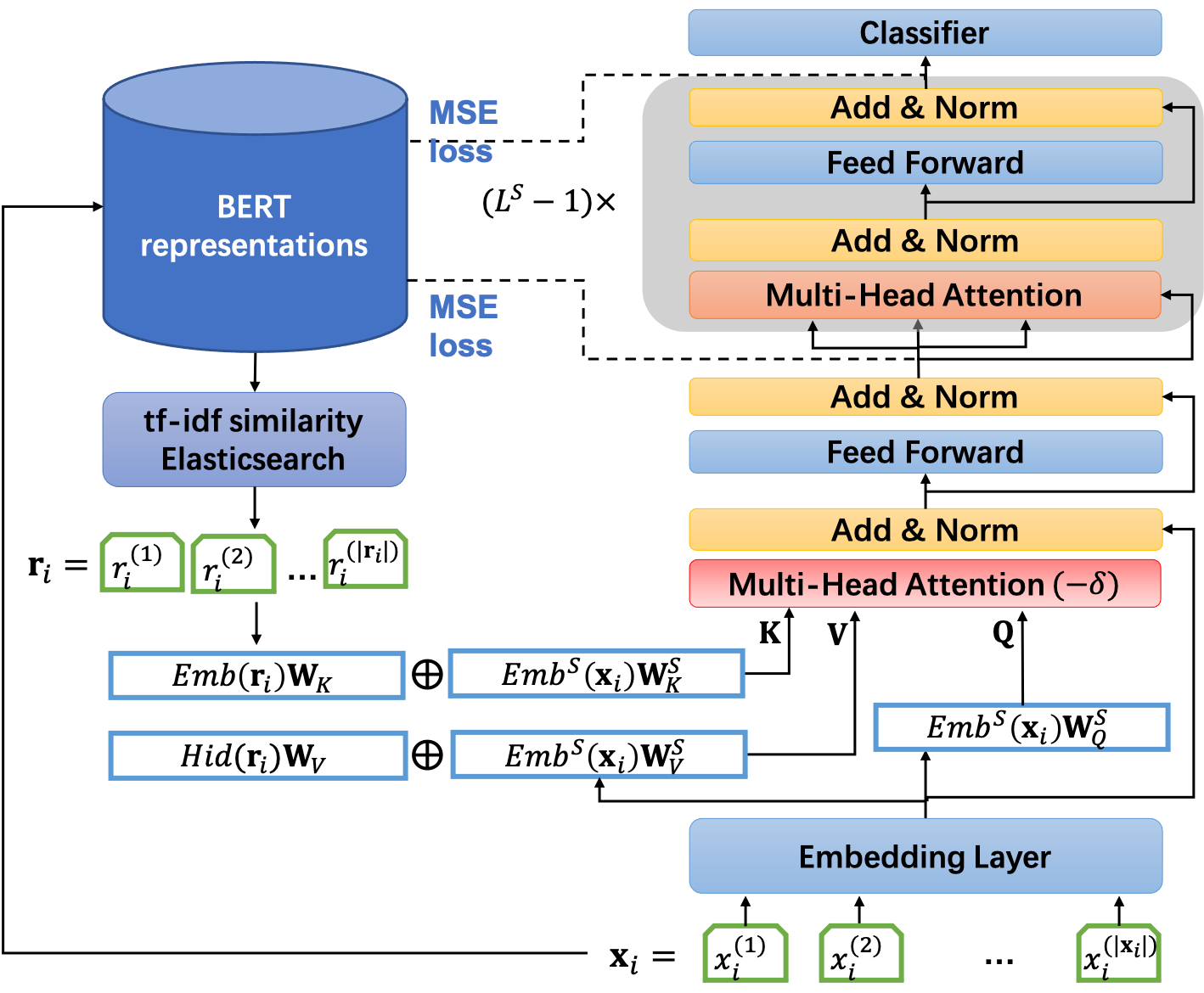}
    \caption{Architecture of RefBERT.}
    \label{fig:model}
\end{figure*}

\subsection{Model Architecture}

Following the setup of TinyBERT~\cite{DBLP:journals/corr/abs-1909-10351}, we break the distillation into two stages: the general distillation (GD) stage and the task distillation (TD) stage.  In the following, we first outline the GD stage.  We first construct a new reference dataset to pair a reference example for each sample in $\gD_{LM}$:
\begin{align}
\gD_R = \{(\rvx, & \rvr)~|~\rvx\in\gD_{LM} \land \rvr = \argmin_{\rvv\in\gD_{LM}\setminus\{\rvx\}} D(\rvv, \rvx)\}.
\end{align}
Hence, $|\gD_R|=|\gD_{LM}|$ and the reference example $\rvr$ in $\gD_R$ is different from $\rvx$ in $\gD_{LM}$, but is the most similar one based on a certain criterion, e.g., containing the most common keywords or following the most similar structure. 

Let the hidden size in the student model be $d_{H^S}$ and the number of Transformer layers be $L^S$.  Usually, we have $d_{H^S}<d_H$ and $L^S<L$ because we want to learn a smaller student model while preserving the teacher's performance.  

As illustrated in Fig.~\ref{fig:model}, given $(\rvx, \rvr)\in\gD_R$, we have $Emb(\rvr)\in\sR^{d_{H^S}}$ and $Hid(\rvr)\in\sR^{d_{H^S}}$ obtained from the teacher model.  We modify the first Transformer layer in the student model to absorb the reference information while keeping the rest layers the same as the teacher model.  More specifically, the first Transformer layer of the student model is computed by 
\begin{align}
    \rvh_1^S &= \gT^S(Emb^S(\rvx), Emb(\rvr), Hid(\rvr))\\\nonumber
    &\defeq\left\{
    \begin{array}{@{}l@{}l@{}}
    \rmA^S &=\MH^S(\rmQ_1^S, \rmK_1^S, \rmV_1^S), \\
    \rmB^S &= \LN(Emb^S(\rvx)+\rmA^S),\\
    \rvh_1^S &= \LN(\FFN(\rmB^S)+\rmB^S),
    \end{array}
    \right.
\end{align}
where
\begin{align}\label{eq:S_Q1}
&\rmQ_h^S\defeq\rmQ_h^S(\rvx) = Emb^S(\rvx)\rmW_Q^{S,h} \\\label{eq:S_K1}
&\rmK_h^S\defeq \rmK_h^S(\rvx, \rvr) = Emb^S(\rvx)\rmW_{K}^{S,h}\Concat Emb(\rvr) \rmW_{K}^h, \\\label{eq:S_V1}
&\rmV_h^S\defeq \rmV_h^S(\rvx, \rvr) = Emb^S(\rvx)\rmW_{V}^{S,h}\Concat Hid(\rvr)\rmW_{V}^h, \\\label{eq:att_S}
& Att_h^S = \frac{\rmQ_h^S\rmK_h^{S\top}}{\sqrt{d_{H^S}}},\\\label{eq:head_S}
& \head_h^S = \left(\softmax(Att_h^S\right) - \delta) \rmV_h^S,\\\label{eq:Multihead_S} 
& \rmA^S = \MH^S(\rmQ_h^S,  \rmK_h^S, \rmV_h^S) = \Concat_{h=1}^{H} \head_h^S. 
\end{align}
To guarantee the compatibility, the number of the attention heads in the student is the same as $H$ in the teach model in Eq.~(\ref{eq:MH}).  
\begin{remark}
We highlight several points about the difference between the student model and the teacher model in the above computation:
\begin{compactitem}
\item At the student's first Transformation layer, the query, key, and value, $\rmQ_1^S$, $\rmK_1^S$, and $\rmV_1^S$, are mapped to the projected space and are slightly different from the original query, key, and value without mapping in the teacher model; see Eq.~(\ref{eq:Tencoder}).  This yields a different computation in the self-attention score as in Eq.~(\ref{eq:att_S}) of the student model contrast to that in Eq.~(\ref{eq:att}) of the teacher model. 
\item Different from the teach model, $\rmK_1^S$ and $\rmV_1^S$ have included the information of both $\rvx_i$ and $\rvr_i$ via concatenation.  In $\rmK_1^S$, the concatenated component is related to the embedding, $Emb(\rvr_i)$, while in $\rmV_1^S$, the component is related to the output in the last Transformer layer, $Hid(\rvr_i)$.  By such setting and the self-attention mechanism, we can absorb the matched information on the input, $Emb(\rvx_i)$ and $Emb(\rvr_i)$, while utilizing the output of $Hid(\rvr_i)$ for the final output. 
\item In Eq.~(\ref{eq:head_S}), we subtract the normalization attention score by a constant $\delta$, where $0<\delta<1$. This is a key setup borrowed from~\cite{DBLP:conf/aistats/WangY20} to make some of the attention weights negative so that gradients would have different signs at back-propagation. It is noted that the original attention score ranges from 0 to 1, where a higher value implies high correlation on the corresponding tokens and a lower value towards un-correlation.  By subtracting a constant in the score, we can place negative attention for non-informative tokens and discard the impact of unrelated parts in the next layer.    

\end{compactitem}
\end{remark}

Hence, for a pair of $(\rvx, \rvr)$, the output of each layer of our RefBERT is $\rvh_l^S \in \sR^{|\rvx|\times H^S}$ and is computed by:
\begin{align}\label{eq:h^S}
    &\rvh_l^S \\\nonumber 
   =& \left\{\begin{array}{@{}l@{}l}
   Emb^S(\rvx),~~ & l=0,  \\ 
   \gT^S(Emb^S(\rvx), Emb(\rvr), Hid(\rvx)),~~ & l=1,  \\
 \gT(\rvh_{l-1}^S),~~ & l=2, \ldots, L^S. 
 \end{array}
 \right.
\end{align} 

\subsection{Model Training}
Following TinyBERT~\cite{DBLP:journals/corr/abs-1909-10351}, we let $n = m(l)$ be a mapping function to indicate the $l$-th layer of student model learns the information from the $n$-layer of teacher model.  Hence, $0=m(0)$ in the embedding layer and $L+1=m(L^S+1)$ in the last Transformer layer.  

In the training, we follow the setup of TinyBERT to apply mean-squared-error (MSE) to distill the information.  That is, both the {\em embedding-layer} and the {\em prediction-layer} are distilled while in the {\em prediction-layer}, the {\em attentions} and {\em hidden states} are distilled.  Hence, we derive the same objective function: 
\begin{align}\label{eq:refbert_loss}
    \gL = \sum_{l=0}^{L^S+1}\lambda_l \gL_{\rm layer}(S_l, T_{m(l)}),
\end{align}
where 
\begin{align}\label{eq:distil_loss}
    & \gL_{\rm layer}(S_l, T_{m(l)}) \\\nonumber
=&  \left\{
\begin{array}{@{~}ll}
\gL_{Emb}(S_0, T_{0}), & l=0\\
\gL_{Hid}(S_l, T_{m(l)})+\gL_{Att}(S_l, T_{m(l)}), & l=1,\ldots, L^S\\
\gL_{P}(S_{L^S+1}, T_{L+1}), & l=L^S+1.
\end{array}
\right. 
\end{align}
In Eq.~(\ref{eq:distil_loss}), the distillation losses consist of:
\begin{compactitem}
\item {\bf Embedding-layer distillation}.  MSE is adopted to penalize the difference between the embeddings in the student and the teacher:
\begin{align}
    \gL_{Emb}(S_0, T_{0}) = \MSE(\rvh_0^S\rmW_e, \rvh_0),
\end{align}
where $\rmW_e\in\sR^{d_{H^S}\times d_H}$ is a weight matrix to linearly map the embedding in the student to the same space as that in the teacher.
\item {\bf Hidden state distillation}.  MSE is adopted to penalize the difference between the hidden states in the student and the teacher: 
\begin{align}
    \gL_{Hid}(S_l, T_{m(l)}) = \MSE(\rvh_l^S\rmW_l, \rvh_{m(l)}),
\end{align}
where $\rmW_l\in\sR^{d_{H^S}\times d_H}$ plays the same role on $\rmW_e$ to linearly map the hidden state in the student to the same space as that in the teacher.
\item {\bf Attention distillation}.  MSE is adopted to penalize the difference between the attention weights in the student and the teacher:
\begin{align}
    \gL_{Att}(S_l, T_{m(l)}) = \frac{1}{H}\sum_{h=1}^{H}\MSE(Att_h^S, Att_h),
\end{align}
where $H$ is the number of attention heads, $Att_h\in\sR^{T\times T}$ is the $h$-th head of student or teacher, $T$ is the length of input text.  It is noted that the attention score is unnormalized and computed as in Eq.~(\ref{eq:att}) or Eq.~(\ref{eq:att_S}) because a faster convergence is verified in TinyBERT.   
\item {\bf Prediction-layer distillation}. The soft cross-entropy loss is adopted on the logits of the student and teacher to mimic the behaviors of intermediate layers:
\begin{align}\nonumber
   & \gL_{P}(S_{L^S+1}, T_{L+1})\!=\!-\softmax(\rvo)\cdot {\rm log}\_\softmax(\rvo^S/t), 
\end{align}
where $\rvo$ and $\rvo^S$ are the logits of the teacher and the students computed from fully-connected networks, respectively.  ${\rm log}\_\softmax(\cdot)$ denotes the log likelihood.  $t$ is a scalar representing the temperature value and usually set to 1.  
\end{compactitem}

\section{Theoretical Analysis}

We provide some theoretical insight into the distillation procedure by examining the mutual information between the student model and the teacher model.  From an information-theoretic perspective, the goal of compressing a large model like BERT is to retain as much information as possible.  That is to maximize the mutual information between the student model and the teacher model.  

Let $(U, V)$ be a pair of random variables with values over the space ${\gU}\times {\gV}$, the mutual information is defined by 
\begin{align}\label{eq:mutual_info}
    I(U; V) = H(U) - H(U|V),
\end{align}
where $H(\cdot)$ denotes the entropy and $H(\cdot|\cdot)$ denote the conditional entropy.  

\noindent\textbf{Justification of the Loss Function.}  Usually, we can apply $U$ to represent the output from the teacher while $V$ as the output of the student.  Since $H(U)$ is fixed, maximizing $I(U; V)$ is equivalent to minimizing $H(U|V)$.  The conditional entropy $H(U|V)$ quantifies the amount of additional information to describe $U$ given $V$.  Since it is difficult to directly compute $H(U|V)$, we derive an upper bound for the conditional entropy during compression.

\begin{theorem}[Upper Bound of Conditional Entropy]\label{thm:ub_cond_entropy}
Let $(U, V)$ be a pair of random variables with values over the space ${\gU}\times {\gV}$, we have
\begin{align}\label{eq:L}
     H(U|V)) \leq \frac{1}{2} \log{\mathbb{E}_{U,V}[(U-\phi(V))^2]} + c,
\end{align}
where $c$ is a constant.
\end{theorem}
\begin{proof}
We introduce a variational distribution $q(U|V) \sim \mathcal{N}(\phi(V), \mathbb{E}_{U,V}[(U - \phi(V))^2])$.  That is, the conditional probability of $U$ given $V$ follows a Gaussian distribution with a diagonal covariance matrix.  Then, we can derive
\begin{align}
     & H(U|V)) \\
     =& -\int_\mathcal{V} \int_\mathcal{U} p(u,v) \log{p(u|v) du dv} \nonumber \\
     =& -\int_\mathcal{V} p(v) \left(\int_\mathcal{U} p(u|v) \log p(u|v) du\right) dv \nonumber \\
    =& -\int_\mathcal{V} p(v) \left(\int_\mathcal{U} p(u|v) \log \left(q(u|v)\cdot \frac{p(u|v)}{q(u|v)}\right) du\right) dv \nonumber \\ 
     =& -\mathbb{E}_V [\mathbb{E}_{U|V} [\log{q(U|V)}]] - KL(p(U|V)||q(U|V)) \nonumber \\
    \leq &  -\mathbb{E}_{U,V} [\log{q(U|V)}]] \nonumber \\
    = & \frac{1}{2} \log{\mathbb{E}_{U,V}[(U-\phi(V))^2]} + c.
\end{align}
In the above, the fourth line holds because $KL(p(U|V)||q(U|V))$ is a constant while the inequality in the fifth line holds because $KL(p(U|V)||q(U|V))$ is non-negative.
\end{proof}
\begin{remark}
Theorem~\ref{thm:ub_cond_entropy} justifies that MSE is an effective surrogate to guarantee information absorption.
\end{remark}

\noindent\textbf{Justification of the Usage of the Reference Sample.}  To see the information flow between the layers, we close up to examine the mutual information between the student and the teacher at each layer.  First, we present the following two theoretical results.

\begin{theorem}[Decrease in Mutual Information]\label{thm:DMI}
For any mapping function $f$, which maps a random variable $U$ to another random variable $V$ through some learning process, we have the following inequality:
\begin{align}\label{eq:ref_ineq}
    I(f(U); V) \leq I(U; V).
\end{align}
\end{theorem} 
The result can be derived by the Data Processing Inequality (DPI)~\cite{10.5555/1146355}.  A similar argument is also stated in~\cite{shwartz2017opening}: in deep learning, the mutual information between the learned representation and the ground truth signal decreases from layer to layer.  
\begin{remark}
Theorem~\ref{thm:DMI} indicates that the difference between the teacher and the student is that the teacher discards the useless information and reserves the most useful information in learning the final representation.  In contrast, the student might discard some useful information in the learning procedure and ends up with a less precise representation.
\end{remark}

\begin{theorem}[Increase of Mutual Information by Reference Samples]\label{thm:IMI}
For random variables $U$, $W$, $V$ where $U$ and $W$ follow the same distribution, we have the following inequality:
\begin{align}\label{eq:diff_I}
    I(U; V) \leq I(U, W; V).
\end{align}
\end{theorem}
\begin{proof}
By the definition in Eq.~(\ref{eq:mutual_info}), we have 
\begin{align*}
  & I(U, W; V) - I(U; V) \\
= & H(U, W) - H(U, W | V) - H(U) + H(U|V) \\
= & H(U, W) - H(U) - H(U, W | V) + H(U|V) \\
= & H(U | W) - H(U|W, V) \\
= & I(U; V | W) \ge 0
\end{align*}
This concludes the results in Eq.~(\ref{eq:diff_I}).
\end{proof}
Note that $I(U, W; V) = I(U; V)$ when $U$ and $W$ are independent.  

\begin{remark}
By deeming $U$ to the teacher's representation, $W$ the  reference sample's representation, and $V$ the student's representation, with Theorem~\ref{thm:IMI}, we conclude that by including any related reference sample during the compression procedure, we can increase the information absorption.  This verifies the effectiveness of our proposed RefBERT.
\end{remark}

\section{Experiments}
\label{sec:exp}
\begin{table*}[htp]
\begin{center}
\caption{The GLUE benchmark} 
\label{tb:GLUE_dataset}
\begin{tabular}{lrrlll}
\hline
 {\bf Corpus} &  $|${\bf Train}$|$ & $|${\bf Dev.}$|$ & {\bf Task} & {\bf Metrics} & {\bf Domain}\\ \hline
 \multicolumn{6}{c}{Single-Sentence Tasks} \\\hline
  CoLA & 8.5k &  1k & acceptability & Matthews corr. & misc.\\
  SST-2  & 67k & 1.8k & sentiment & acc. & movie reviews\\\hline
  \multicolumn{6}{c}{Similarity and Paraphrase Tasks} \\\hline
MRPC & 3.7k &  1.7k & paraphrase & acc./F1 & news\\
STS-B & 7k & 1.4k & sentence similarity & Pearson/Spearman corr. & misc.\\
QQP & 364k & 391k & paraphrase & acc./F1 & social QA questions\\\hline
\multicolumn{6}{c}{Inference Tasks}\\\hline
MNLI & 393k & 20k & NLI & matched acc./mismatched acc. & misc.\\
QNLI & 105k &  5.4k & QA/NLI & acc. & Wikipedia  \\
RTE & 2.5k & 3k & NLI & acc.  & news, Wikipedia \\
WNLI & 634 & 146 & coreference/NLI & acc. & fiction books \\\hline
\end{tabular}
\end{center}
\end{table*}   
In this section, we evaluate RefBERT on the General Language Understanding Evaluation (GLUE) benchmark and show the effectiveness of our RefBERT on utilizing the reference samples.

\begin{table*}[t]
\caption{Results are evaluated on the test set of GLUE official benchmark.  All models are learned in a single-task manner.  Note that TinyBERT-DA denotes TinyBERT with data augmentation.  TinyBERT is TinyBERT without data augmentation and is evaluated on the dev set of GLUE.  To fair comparison, RefBERT is also evaluated on the dev set of GLUE.  Mcc. refers to Matthews correlation and Pear./Spea. refer to Pearson/Spearman.
}
\label{tb:GLUE_rs}
\begin{center}
\begin{tabular}{l|ccccccccc|c}
\hline 
Model & CoLA & MNLI-m & MNLI-mm & MRPC & QNLI & QQP & RTE  & SST-2 & STS-B & \multirow{2}{*}{\bf Avg.} \\
      & Mcc. &  Acc.  &   Acc.  &  F1   & Acc. & F1. & Acc. & Acc.  & Pear./Spea. & 
\\ \hline 
$\BERTBASE$ & 52.1 & 84.6 & 83.4 & 88.9 & 90.5 & 71.2 & 66.4 & 93.5 & 85.8 & 79.6/77.3 \\\hline 
DistillBERT & 32.8 & 78.9 & 78.0 & 82.4 & 85.2 & 68.5 & 54.1 & 91.4 & 76.1 & 71.9/68.9  \\
TinyBERT-DA  & 43.3 & 82.5 & 81.8 & 86.4 & 87.7 & 71.3 & 62.9 & 92.6 & 79.9 & 76.5/73.5 \\
TinyBERT & 29.8 & 80.5 & 81.0 & 82.4 & - & - & - & - & - & ~~-~~/68.4 \\\hline 
RefBERT & 47.9 & 80.9 & 80.3 & 86.9 & 87.3 & 61.6 & 61.7 & 92.9 & 75.0 & 75.1/74.0
\\\hline
\end{tabular}
\end{center}
\end{table*}
\subsection{Model Settings}
The code is written in PyTorch.  To provide a fair comparison, we apply the same setting as the TinyBERT.  That is, the number of layers $L^S=4$, the hidden size $d_{H^S} = 312$, the feed-forward/filter size $d_{f^S} = 1,200$ and the head number $H=12$.  This yields a total of 14.8M parameters, where the additional parameters come from the projection matrices of $\rmW_Q^S$, $\rmW_K^S$, and $\rmW_V^S$ in Eq.~(\ref{eq:S_Q1})-Eq.~(\ref{eq:S_V1}).  $\BERTBASE$ is adopted as the teacher model and consists of 109M parameters by the default setting: the number of layers $L=12$, the hidden size $d_{H} = 768$, the feed-forward/filter size $d_{f} = 3,072$ and the head number $H=12$.  The same as TinyBERT, we adopt $m(l)=3l$ as the layer mapping function to learn from every 3 layers of $\BERTBASE$.  The weight loss $\lambda$ at each layer in Eq.~(\ref{eq:refbert_loss}) is set to 1 due to good performance. 


\noindent{\bf General distillation.}  We use the English Wikipedia (2,500M words) as the $\gD_{LM}$ dataset and follow the same pre-processing as in~\cite{DBLP:conf/naacl/DevlinCLT19}.  Each input sentence in $\gD_{LM}$ is paired with a reference sentence by BM25 in Elastic Search~\footnote{\url{https://elasticsearch-py.readthedocs.io/en/v7.10.1/}}.  The model of RefBERT is then trained on 8 16GB V100 GPUs for approximately 100 hours.  

{
\noindent{\bf Task-specific Distillation.} The same as general distillation, we select the reference samples by BM25 in Elastic Search from the English Wikipedia. We then fine-tune RefBERT on the downstream tasks with the teacher's representations on the reference samples.  The learning rate is tuned from 1e-5 to 1e-4 with the step of 1e-5 to seek the best performance on the development sets of the corresponding tasks in GLUE. 

}


\subsection{Dataset}
The GLUE benchmark~\cite{DBLP:conf/iclr/WangSMHLB19} is a collection
of 9 natural language understanding tasks as listed in Table~\ref{tb:GLUE_dataset}:
\begin{compactitem}
\item {\bf CoLA.} The Corpus of Linguistic Acceptability is a task to predict whether an English sentence is a grammatically correct one and evaluated by Matthews correlation coefficient~\cite{DBLP:journals/tacl/WarstadtSB19}.
\item {\bf MNLI.} Multi-Genre Natural Language Inference is a large-scale, crowd-sourced entailment classification task and evaluated by the matched and mismatched accuracy~\cite{DBLP:conf/naacl/WilliamsNB18}.  Given a pair of (premise, hypothesis), the task is to predict whether the hypothesis is an entailment, contradiction, or neutral with respect to the premise.
\item {\bf MRPC.} Microsoft Research Paraphrase Corpus is a paraphrase identification dataset evaluated by F1 score~\cite{DBLP:conf/acl-iwp/DolanB05}.  The task is to identify if two sentences are paraphrases of each other. 
\item {\bf QNLI.} Question Natural Language Inference is a version of the Stanford Question Answering Dataset which has been converted to a binary sentence pair classification task and is evaluated by accuracy.  Given a pair of (question, context), the task is to determine whether the context contains the answer to the question. 
\item {\bf QQP.}  Quora Question Pairs is a collection of question pairs from the website Quora. The task is to determine whether two questions are semantically equivalent and is evaluated by the F1 score.
\item {\bf RTE.} Recognizing Textual Entailment is a binary entailment task with a small training dataset and is evaluated by the accuracy~\cite{DBLP:conf/tac/BentivogliMDDG09}.
\item {\bf SST-2.} The Stanford Sentiment Treebank is a binary single-sentence classification task, where the goal is to predict the sentiment of movie reviews and is evaluated by the accuracy~\cite{DBLP:conf/emnlp/PenningtonSM14}. 
\item {\bf STS-B.}  The Semantic Textual Similarity Benchmark is a collection of sentence pairs drawn from news headlines and many other domains. The task aims to evaluate how similar two pieces of texts are by a score from 1 to 5 and is evaluated by the Pearson correlation coefficient~\cite{DBLP:journals/corr/abs-1708-00055}.
\end{compactitem}
We follow the standard splitting to conduct the experiments and submit the prediction results to the GLUE benchmark  system~\footnote{\url{https://gluebenchmark.com/}} to evaluation the performance on the test sets.

\begin{table}[t]
\caption{The model sizes and inference time for the  baselines and RefBERT.  The number of layers does not include the embedding and prediction layers. 
}
\label{tb:Inference_time}
\begin{center}
\begin{tabular}{@{~}l@{~}|@{~}c@{~~}c@{~~}c@{~~}c@{~~}c@{~}}
\hline 
Model & Layers & Hidden & Feed-forward &  Model & Inference \\
& & Size & Size & Size & Time (s) \\\hline
$\BERTBASE$ & 12 & 768 & 3072 & 109M ($\times$1.0) & 190 ($\times$1.0) \\\hline 
DistillBERT & 4 & 768 & 3072 & 52.2M ($\times$2.1) & 64.1 ($\times$3.0)\\
TinyBERT & 4 & 312 & 1200 & 14.5M ($\times$7.5) & 20.1  ($\times$9.5)\\
RefBERT & 4 & 312 & 1200 & 14.8M ($\times$7.4) & 20.1 ($\times$9.5) \\\hline
\end{tabular}
\end{center}
\end{table}

\subsection{Results}
Table~\ref{tb:GLUE_rs} reports the comparison results between our RefBERT and baselines, $\BERTBASE$, DistilBERT, TinyBERT with Data Augmentation (TinyBERT-DA), and the vanilla TinyBERT.  The results of baselines are copied from those reported  in TinyBERT~\cite{DBLP:journals/corr/abs-1909-10351} for reference.  The results show that RefBERT can attain competitive performance on all tasks in GLUE benchmark: 
\begin{compactitem}
\item RefBERT gains 7 better results than  DistilBERT out of the 9 tasks and obtains a prominent improvement of 4.4\% on average.  
\item RefBERT significantly outperforms the vanilla TinyBERT in all compared tasks and attains a large improvement of 8.1\% on average of the four compared tasks, CoLA, MNLI-m, MNLI-mm, and MRPC.  
\item RefBERT can even obtain better performance on CoLA, MRPC, and STS-2 compared to TinyBERT-DA and yields 98.2\% performance of TinyBERT-DA.  
\item Overall, the average performance of RefBERT is at least 94\% of $\BERTBASE$.  RefBERT attains relative lower performance on the task of QQP.  We conjecture that the tokens in the reference samples in QQP may be too similar to those in the evaluated samples and make confusion the prediction.  
\end{compactitem}

In terms of the inference time reported in Table~\ref{tb:Inference_time}, we observe that RefBERT contains slightly larger model size, around 300K parameters, than TinyBERT.  However, the inference time is negligible.  Compared with the teacher $\BERTBASE$, RefBERT is 7.4x smaller and 9.5x faster while achieving competitive  performance.  The results show that RefBERT is a promising surrogate for the recently-developed BERT distillation models.   



\section{Conclusion}
In this paper, we propose a new knowledge distillation method, namely RefBERT, to distill BERT by utilizing the teacher's representations on the reference samples.  By including the references' word embeddings and the teacher's final layer  representations in the corresponding key and value while amplifying the self-attention effect of irrelevant components in the first layer, we can make RefBERT absorb the teacher's knowledge on the reference samples and strengthen the information interaction effectively.  More importantly, we provide theoretical justification on selecting the mean-square-error loss function and prove that including reference samples indeed can increase the mutual information of distillation.  Our experimental evaluation shows that RefBERT can beat the vanilla TinyBERT over 8.1\% and achieves more than 94\% of the performance of $\BERTBASE$ on the GLUE benchmark.  Meanwhile, RefBERT is 7.4x smaller and 9.5x faster on inference than $\BERTBASE$.

Several research problems are worthy of further exploration.  First, we would like to explore more ways to reduce the model size of RefBERT while maintaining the same performance.  Second, it would be promising to investigate more effective mechanisms to transfer the knowledge from wider and deeper teachers, e.g., BERT-large, to a smaller student via the reference mechanism.  Third, other speed-up methods, e.g., quantization, pruning, and even hardware acceleration, can be attempted to resolve the computation overhead on the large pre-trained language models.

\bibliography{reference} 
\bibliographystyle{plain}

\end{document}